\documentclass{article}
\usepackage{spconf,amsmath,epsfig}

\usepackage{amsmath}
\usepackage{amsthm}
\usepackage{booktabs}
\usepackage{algorithm}
\usepackage{algorithmic}
\usepackage{multirow}
\usepackage{amsfonts} 

\usepackage{setspace} 
\usepackage[dvipsnames, svgnames, x11names]{xcolor} 
\usepackage{tikz}
\usetikzlibrary{arrows}
\usepackage{graphicx}
\usepackage{subfigure}
\usetikzlibrary{shapes}

\usepackage{marvosym}

\let\OLDthebibliography\thebibliography
\renewcommand\thebibliography[1]{
  \OLDthebibliography{#1}
  \setlength{\parskip}{0pt}
  \setlength{\itemsep}{0pt plus 0.3ex}
}

\begin{document}\sloppy
\begin{spacing}{0.885}
\def\x{{\mathbf x}}
\def\L{{\cal L}}

\title{A Result based Portable Framework for Spoken Language Understanding}
%
\name{Lizhi Cheng$^{1,2}$, \Letter Weijai Jia$^2$, \Letter Wenmian Yang$^1$}
\address{
$^1$Department of Computer Science and Engineering, Shanghai Jiao Tong University, Shanghai, China\\
$^2$BNU-UIC Institute of Artificial Intelligence and Future Networks
Beijing Normal University (BNU Zhuhai)\\
Guangdong Key Lab of AI and Multi-Modal Data Processing, BNU-HKBU United International College\\
Zhuhai, Guangdong, PR China\\
clz19960630@sjtu.edu.cn, jiawj@sjtu.edu.cn
}

\maketitle

\begin{abstract}
Spoken language understanding (SLU), which is a core component of the task-oriented dialogue system, has made substantial progress in the research of single-turn dialogue. However, the performance in multi-turn dialogue is still not satisfactory in the sense that the existing multi-turn SLU methods have low portability and compatibility for other single-turn SLU models. Further, existing multi-turn SLU methods do not exploit the historical predicted results when predicting the current utterance, which wastes helpful information. To gap those shortcomings, in this paper, we propose a novel Result-based Portable Framework for SLU (RPFSLU). RPFSLU allows most existing single-turn SLU models to obtain the contextual information from multi-turn dialogues and takes full advantage of predicted results in the dialogue history during the current prediction. Experimental results on the public dataset KVRET have shown that all SLU models in baselines acquire enhancement by RPFSLU on multi-turn SLU tasks.

\end{abstract}
\begin{keywords}
Spoken language understanding, dialogue systems, multimedia application, multi-task learning
\end{keywords}
\section{Introduction}
\label{intro}
\begin{figure}[t]
\centering
\includegraphics[width=0.8\columnwidth]{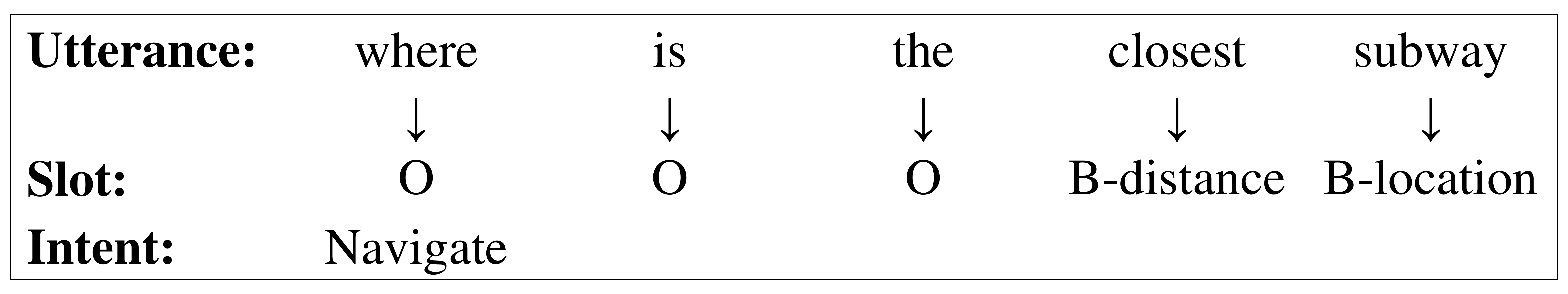}
\caption{An example of SLU.}
\label{fig:example}
\end{figure}

As a popular multimedia application, task-oriented dialogue systems have been widely applied in intelligent voice assistant, e.g., Apple Siri, where Spoken Language Understanding (SLU) plays a critical role. Given an utterance expressed in natural language, SLU aims to form a semantic frame that captures the semantics of user utterances or queries. Generally, SLU consists of two main subtasks, i.e., Intent Detection (ID) and Slot Filling (SF) \cite{SLU2011}. Intent detection is a semantic classification problem that analyzes the semantics of the utterance on sentence-level, and slot filling is a sequence labeling task that works on word-level (token level) \cite{bi-dictional2019}. A simple example of SLU is shown in Figure \ref{fig:example}.

Traditional AI voice assistant mainly makes single-turn communication with the user. However, in real application scenarios, people often need to make a multi-turn dialogue with the voice assistant to achieve their demand. 
Therefore, the multi-turn dialogue-based SLU task is increasingly crucial. 
Although some studies \cite{End-to-end2016,sequentialSlu2017,memory2019} have proposed some approaches to deal with the multi-turn dialogue task by combining contextual information from dialogue history, their basic SLU models are not state-of-the-art compared to the latest work in single-turn SLU \cite{Slot-gated2018,JointCapsule2019,bi-dictional2019,qin2019stack}.
Moreover, since their basic SLU models are coupled with other modules in their inner structure, it is also difficult to replace them with state-of-the-art single-turn SLU models.
The above problem makes the multi-turn SLU approaches not benefit from the latest work in single-turn SLU directly.
Therefore, designing a portable multi-turn SLU framework that can utilize most existing single-turn SLU models as basic models without changing their inner structure is a significant issue.

In addition, existing multi-turn SLU approaches only consider the utterances in the dialogue history when obtaining contextual information, whereas they largely ignore the previously predicted SLU results of these utterances. However, these predicted results also contain useful semantic information and may affect current predicting.  An empirical explanation is that the topic of the current utterance is usually related to the previous utterances in multi-turn dialogue, so does its SLU results. Thus, the historical results are worthy of the reference for the current prediction.

Moreover, in every single turn, the prediction results of ID and SF are also correlative and have direct interaction. For example, in Figure \ref{fig:example}, if the intent is detected as \texttt{Navigate}, then the token ``subway" is more likely to be recognized as the slot \texttt{location}. But if the intent is detected as \texttt{BookingRestaurant}, then the token ``subway" is more likely to be filled as \texttt{ResturantName}. Conversely, when we have filled some slot labels e.g., \texttt{ResturantName}, \texttt{DinningTime}, but find the intent is \texttt{GetWeather}, it is obvious that the intent label is incorrect.
In the case above, the result of ID can affect the labeling of SF, while the results of the SF can verify the correctness of ID, which shows that it is precious to incorporate prediction results into the model.
Currently, only stack propagation strategy \cite{qin2019stack} try to directly utilize the probability distributions of predicted ID results to support SF predicting. However, their method not only ignores the semantics of results but also completely ignores the effect of SF results, which is ineffective.
Therefore, how to effectively leverage the semantic information contained in both predicted ID and SF results of SLU tasks is also a meaningful issue.



To solve the above issues, in this paper, we propose a Result-based Portable Framework for SLU (RPFSLU).  RPFSLU allows most existing single-turn SLU models to obtain contextual information from multi-turn dialogues and takes full advantage of predicted results in the predicting process. In RPFSLU, existed single-turn SLU models (so-called basic model) work as a black-box that only needs to provide prediction results. So there is no need to understand or change their inner structure.
Generally, RPFSLU comprises two parts, i.e., Dialogue History Representation (DHR) and Result-based Bi-Feedback Network (RBFN).
DHR aims to obtain contextual information from both historical utterances and historical predicted results. RBFN aims to incorporate the predicted results of ID and SF in the network and make a more accurate prediction by utilizing the semantic information contained in results. More specifically, DHR inputs the current utterance with its dialogue history, including historical utterances and predicted results, and outputs an embedded sequence containing contextual information. RBFN contains a two-round prediction process. In practice, we first obtain the first-round prediction results of ID and SF from the basic model (any single-turn SLU model) and embed the results to the latent state vector. Subsequently, we merge the latent state vector with the word embedding of the utterance and resend the merged embedding into the basic model to predict the second-round results. Finally, we combine the results from two rounds and output the final SLU results. 

The main contributions of this paper are presented as follows:
\begin{enumerate}
\item We propose a portable framework that allows most existing single-turn SLU models to take full advantage of contextual information and be able to deal with multi-turn SLU tasks without changing their inner structure.
\item We focus on the impact of prediction results on SLU tasks. We efficiently utilize both the predicted results from dialogue history and the interaction between ID and SF in every single turn to improve the final performance of existing SLU models without using any extra labeled data.
\item We present extensive experiments demonstrating the beneﬁt of RPFSLU. Experimental results show that all SLU models in baselines improve their performance when using RPFSLU. 
\end{enumerate}


\section{Related Work}
In SLU, intent detection is usually seen as a semantic classification problem to predict the intent label, and slot filling is mainly regarded as a sequence labeling task.
With recent developments in deep neural networks, many methods \cite{qin2021survey,yao2014lstm,kurata2016lstm} have been proposed to solve these two tasks. Traditionally, pipeline approaches are utilized to manage the two mentioned tasks separately. These kinds of methods usually suffer from error propagation due to their independent models.
Motivated by this problem, the joint models \cite{JointModel2016,chen2016joint,attention-basedRnn2016} are developed for solving intent detection and slot filling tasks together.
Besides, some work \cite{Multi-domainJoint2016,memory2019,Multi-TaskNetworks2019,qin2020co,qin2020dcr,qin2021knowing}, tries to enhance the performance via multi-task learning.
These joint models or multi-task learning methods link the two tasks implicitly via applying a joint loss function.
Considering that slot and intent have a strong relationship, slot-gated model \cite{Slot-gated2018} focuses on learning the relationship between intent and slot attention vectors to obtain better semantic frame results by the global optimization. 
Moreover, in some works \cite{bi-dictional2019,liu2019cm}, the interrelated connections for the two tasks can be established via a collaborative memory network. A hierarchical capsule neural network structure \cite{JointCapsule2019} is also proposed to encapsulate the hierarchical relationship among utterance, slot, and intent.
BERT \cite{BERT} is also utilized in some SLU models \cite{chen2019bert}.
The state-of-art joint model Stack-Propagation \cite{qin2019stack} utilized result information of ID to guide SF task but ignores the impact of SF results on the ID task.

Compared with their work, our model focus on the direct interaction of the prediction results between two tasks, which makes the impact between two tasks more efficient.


\section{Method}


\subsection{Problem Formulation}
\label{3.1}
The input of multi-turn SLU tasks is a sequence of user utterances $\textbf{U}=\{ \textbf{u}_1,\textbf{u}_2,...,\textbf{u}_{n} \}$, where $n$ represents the total number of utterances.
Particularly, $n = 1$ implies the input has no dialogue history.
For any $\textbf{u}_t \in \textbf{U}$, $\textbf{u}_t=\{ x_1,x_2,...x_{k} \}$ is a token sequence, where $k$ represents the number of tokens in utterance $\textbf{u}_t$.


Given $\textbf{U}$ as input, our task is composed of two subtasks, i.e., ID and SF.
ID is a semantic classification task to predict the intent label for each utterance in $\textbf{U}$, and SF is a sequence labeling task to give each token in the utterance a slot label.

\subsection{Overview}
\label{3.2}

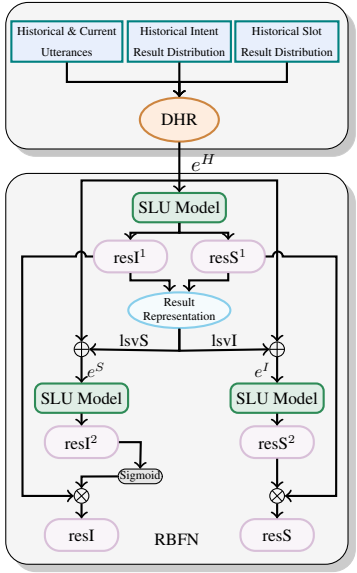
\begin{figure}[t]
\centering
\tikzstyle{background1} = [rectangle, rounded corners=3mm,minimum width = 7cm, minimum height = 3cm, text centered, draw = black, fill = gray!8]
\tikzstyle{background2} = [rectangle, rounded corners=3mm,minimum width = 7cm, minimum height = 8cm, text centered, draw = black, fill = gray!8]
\tikzstyle{shape1} = [rectangle,minimum width = 2cm, minimum height = 0.8cm, text centered, draw = Teal, fill = CornflowerBlue!15,thick]
\tikzstyle{dhr} = [ellipse,inner sep=2mm, text centered, draw = Peru,fill =Peach!15,thick]
\tikzstyle{SLU_model} = [rectangle,rounded corners=1mm,minimum width = 1.6cm, minimum height = 0.6cm, text centered, draw = SeaGreen, fill = SeaGreen!15,thick]
\tikzstyle{res} = [rectangle,rounded corners=2mm,minimum width = 1.5cm, minimum height = 0.65cm, text centered, draw = Thistle, fill = Thistle!15,thick]
\tikzstyle{RR} = [ellipse,inner sep=0, text centered, draw = SkyBlue, fill = SkyBlue!8,thick]
\tikzstyle{addition} = [circle,minimum size =0.3cm,inner sep=0pt, text centered, draw = black]
\tikzstyle{sigmoid} = [rectangle,inner sep=1pt,rounded corners=1mm,minimum width = 0.6cm, minimum height = 0.3cm, text centered, draw = black, fill = gray!20]

\begin{tikzpicture}[node distance = 0cm, scale=0.65]
\tikzstyle{every node}=[scale=0.65]
\node(background1_shadow)[background1,fill=gray!45,draw=gray!40]{};
\node(background1)[background1,below of=background1_shadow,xshift=-0.15cm,yshift=0.15cm]{};
\node(background2_shadow)[background2,below of=background1_shadow,yshift=-6cm,fill=gray!40,draw=gray!40]{};
\node(background2)[background2,below of = background1,yshift=-6cm]{};
\node(hird)[shape1,below of = background1,xshift=0.05cm,yshift=0.7cm,align=center]{\scriptsize Historical Intent\\ \scriptsize Result Distribution};
\node(hsrd)[shape1,right of = hird,xshift=2.2cm,yshift=0cm,align=center]{\scriptsize Historical Slot\\ \scriptsize Result Distribution};
\node(hcu)[shape1,left of = hird,xshift=-2.3cm,yshift=0cm,align=center]{\scriptsize Historical \& Current\\ \scriptsize Utterances};
\node(dhr)[dhr,rounded corners=4mm,below of = hird,xshift=0cm,yshift=-1.6cm,align=center]{DHR};
\node(SLU_1)[SLU_model,below of=dhr,yshift=-1.8cm]{SLU Model};
\node(resi1)[res,below of=SLU_1,xshift=-1cm,yshift=-1cm,align=center]{resI$^{1}$};
\node(ress1)[res,below of=SLU_1,xshift=1cm,yshift=-1cm,align=center]{resS$^{1}$};
\node(rr)[RR,below of=SLU_1,xshift=0cm,yshift=-2.1cm,align=center]{\scriptsize Result \\[-0.8ex] \scriptsize Representation };
\node(SLU_2)[SLU_model,below of=rr,xshift=2cm,yshift=-1.8cm]{SLU Model};
\node(SLU_3)[SLU_model,below of=rr,xshift=-2cm,yshift=-1.8cm]{SLU Model};
\node(resi2)[res,below of=SLU_3,xshift=0cm,yshift=-0.9cm,align=center]{resI$^{2}$};
\node(ress2)[res,below of=SLU_2,xshift=0cm,yshift=-0.9cm,align=center]{resS$^{2}$};
\node(resi3)[res,below of=resi2,xshift=0cm,yshift=-1.9cm,align=center]{resI};
\node(ress3)[res,below of=ress2,xshift=0cm,yshift=-1.9cm,align=center]{resS};
\node(addition1)[addition,above of =SLU_2,yshift=1cm]{};
\draw[-]([xshift=-0.1mm,thick]addition1.east)--([xshift=0.1mm]addition1.west);
\draw[-]([yshift=0.1mm,thick]addition1.south)--([yshift=-0.1mm]addition1.north);
\node(addition2)[addition,above of =SLU_3,yshift=1cm]{};
\draw[-]([xshift=-0.1mm,thick]addition2.east)--([xshift=0.1mm]addition2.west);
\draw[-]([yshift=0.1mm,thick]addition2.south)--([yshift=-0.1mm]addition2.north);
\node(mul1)[addition,above of =ress3,yshift=0.8cm]{};
\draw[-]([xshift=-0.1mm,thick]mul1.south east)--([xshift=0.1mm]mul1.north west);
\draw[-]([yshift=0.1mm,thick]mul1.south west)--([yshift=-0.1mm]mul1.north east);
\node(mul2)[addition,above of =resi3,yshift=0.8cm]{};
\draw[-]([xshift=-0.1mm,thick]mul2.south east)--([xshift=0.1mm]mul2.north west);
\draw[-]([yshift=0.1mm,thick]mul2.south west)--([yshift=-0.1mm]mul2.north east);
\node(sigmoid)[sigmoid,right of=mul2,xshift=1.2cm,yshift=0.4cm,align=center]{\scriptsize Sigmoid};

\draw[->,thick](hird.south) -- (dhr.north);
\draw[->,thick](hsrd.south) --([yshift=-0.4cm]hsrd.south)--([yshift=-0.4cm]hird.south)-- (dhr.north);
\draw[->,thick](hcu.south) --([yshift=-0.4cm]hcu.south)--([yshift=-0.4cm]hird.south)-- (dhr.north);
\draw[->,thick](dhr.south) --node[xshift=0.5cm,yshift=0.1cm,scale=1.2]{$e^{H}$} (SLU_1);
\draw[->,thick](SLU_1.south) --([yshift=-0.2cm]SLU_1.south)--([xshift=-1cm,yshift=-0.2cm]SLU_1.south)-- (resi1.north);
\draw[->,thick](SLU_1.south) --([yshift=-0.2cm]SLU_1.south)--([xshift=1cm,yshift=-0.2cm]SLU_1.south)-- (ress1.north);
\draw[->,thick](resi1.south) --([yshift=-0.15cm]resi1.south)--([xshift=0.8cm,yshift=-0.15cm]resi1.south)-- ([xshift=-0.2cm,yshift=0cm]rr.north);
\draw[->,thick](ress1.south) --([yshift=-0.15cm]ress1.south)--([xshift=-0.8cm,yshift=-0.15cm]ress1.south)-- ([xshift=0.2cm,yshift=0cm]rr.north);
\draw[->,thick](rr.south) --([yshift=-0.45cm]rr.south)--node[yshift=0.2cm]{lsvI} (addition1.west);
\draw[->,thick](rr.south) --([yshift=-0.45cm]rr.south)--node[yshift=0.2cm]{lsvS} (addition2.east);
\draw[->,thick](addition1.south) --node[xshift=-0.3cm]{ $e^{I}$} (SLU_2.north);
\draw[->,thick](addition2.south) --node[xshift=0.3cm]{ $e^{S}$} (SLU_3.north);
\draw[->,thick](SLU_3.south) --(resi2.north);
\draw[->,thick](SLU_2.south) --(ress2.north);
\draw[->,thick](ress2.south) --(mul1.north);
\draw[->,thick](mul1.south) --(ress3.north);
\draw[->,thick](mul2.south) --(resi3.north);
\draw[->,thick](resi2.east) --([xshift=0.4475cm]resi2.east)--(sigmoid.north);
\draw[->,thick](sigmoid.west) --([yshift=0.24cm]mul2.north)--(mul2.north);
\draw[->,thick]([yshift=0.24cm]SLU_1.north) --([yshift=3.3cm]addition1.north)--(addition1.north);
\draw[->,thick]([yshift=0.24cm]SLU_1.north) --([yshift=3.3cm]addition2.north)--(addition2.north);
\draw[->,thick](ress1.east) --([xshift=0.15cm]ress1.east) arc (180:0:0.1cm)--([xshift=1.45cm]ress1.east)--([xshift=1.05cm]mul1.east)--(mul1.east);
\draw[->,thick](resi1.west) --([xshift=-0.15cm]resi1.west) arc (0:180:0.1cm)--([xshift=-1.45cm]resi1.west)--([xshift=-1.05cm]mul2.west)--(mul2.west);
\node()[below of=background2,xshift=0cm,yshift=-3.5cm]{RBFN};
\end{tikzpicture}
\caption{General framework.}
\label{fig:model}
\end{figure}

In this section, we describe the overview of RPFSLU (shown in Figure. \ref{fig:model}) and introduce the relationship between DHR and RBFN in the framework.

RPFSLU aims to make existing single-turn SLU models (so-called basic model) available in multi-turn SLU tasks and take full advantage of prediction results in predicting process. Since the basic model works as a black-box in RPFSLU that only needs to provide prediction results. There is no need to understand or change their inner structure. Thus, RPFSLU can benefit most existing SLU models (e.g., Joint Seq.\cite{JointModel2016}, Attention BiRNN \cite{attention-basedRnn2016}, Bi-Model \cite{birnn2018}, Slot-gated \cite{Slot-gated2018}, SF-ID \cite{bi-dictional2019}, Stack-Propagation \cite{qin2019stack}, etc.).

In a nutshell, RPFSLU is composed of two parts: DHR and RBFN.
DHR works at the very beginning of the whole network and aims to gain semantic information contained in both utterance and prediction results from dialogue history and offering these contextual information to the basic model. And RBFN aims to incorporate semantic information contained in prediction results to the basic model and enhance the performance during every single turn. 

More specifically, the input of DHR is composed of historical utterances $\{\textbf{u}_1,...,\textbf{u}_{T-1}\}$, historical prediction results of ID $\textbf{RESI}=\{\textbf{resI}_1,...,\textbf{resI}_{T-1}\}$, historical prediction results of SF $\textbf{RESS}=\{ \textbf{resS}_1,...,\textbf{resS}_{T-1}\}$, and the current utterance $\textbf{u}_T$, where $\textbf{resI}_t \in \mathbb{R}^{d_i}$, $d_i$ is the category of the intent label, $\textbf{resS}_t = \{ \textbf{s}_1, ... , \textbf{s}_{k}\}$, $s_j \in \mathbb{R}^{d_{s}}$ is the SF result of token $j$, and $d_s$ is the category of the slot label. And the output of DHR is a dialogue-history-related embedding sequence $\textbf{e}^H \in {\mathbb{R}}^{d_{w}}$, where $d_w$ is the dimension of the embedding layer.
Particularly, when dialogue history is none (i.e., single-turn dialogue), DHR works as a normal embedding layer.


As for RBFN, the input is a word embedding sequence, and the output is the final results of ID $\textbf{resI}$ and SF $\textbf{resS}$. As shown in Figure \ref{fig:model}, RBFN contains two-round predictions.
In the first round prediction, the RBFN employs a basic model to obtain the results of ID $\textbf{resI}^1$, and SF $\textbf{resS}^1$.
Then RBFN merges word embedding $\textbf{e}^H$ (received from DHR) with result embeddings of
$\textbf{resI}^1$ and $\textbf{resS}^1$ (calculated by the result representation mechanism that will introduce in the next section), respectively. Meanwhile, we obtain the embedding sequence with intent information $\textbf{e}^I \in {\mathbb{R}}^{d_{w}}$ and the embedding sequence with slot information $\textbf{e}^S \in {\mathbb{R}}^{d_{w}}$.
In the second round prediction, the basic model utilizes $\textbf{e}^S $ as input to obtain the second round ID result $\textbf{resI}^2$ and $\textbf{e}^I$ as input to get the second round SF results $\textbf{resS}^2$.   
We finally merge the prediction results of two rounds and obtain the final ID result $\textbf{resI}$ and SF results $\textbf{resS}$.
In this paper, the basic model utilized in two rounds is the same.

Next, we will introduce the detail of our framework in the following sections.



\subsection{Result Representation Mechanism}
In this section, we introduce result representation mechanism in detail, which is very useful in both DHR and RBFN. 

In SLU, each category of a prediction result has a specific meaning, hence contains essential semantic information. To utilize the semantics of the results efficiently, we design a result representation mechanism that aims to represent the distribution of the prediction results by a specific latent state vector.

Specifically, inspired by \cite{yang2019}, we first employ two latent state matrices (embedding layers), i.e., $\textbf{S}^{I} \in \mathbb{R}^{d_I \times d_{i}}$ and $\textbf{S}^{S} \in \mathbb{R}^{d_S \times d_{s}}$ to express the latent states of the ID and SF results, where $d_{I}$ is the dimension of intent latent states, and $d_{S}$ is the dimension of slot latent states.

Then, for ID, we combine the result distribution $\textbf{resI}$ with latent state matrix $\textbf{S}^{I}$,and obtain the result-based latent state vector $\textbf{lsvI} \in {\mathbb{R}}^{d_{I}}$ by
\begin{equation}
    \textbf{lsvI} =  {\textbf{S}}^{I} \cdot \textbf{resI}
\label{eq:lsvI}
\end{equation}

Similarly as ID, for the SF result $\textbf{s}_{j} $ of token $j$ given by $\textbf{resS} = \{ \textbf{s}_1, ... , \textbf{s}_{k} \}$, we also obtain the result-based latent state vector $\textbf{ls}_j$ by
\begin{align} 
    &\textbf{ls}_j= \textbf{S}^{S} \cdot \textbf{s}_j 
\label{eq:ls}
\end{align}

Moreover, since SF returns a sequence of latent state vectors, we further design an attention mechanism to calculate the weighted average of the sequence and obtain an utterance-level latent state vector $\textbf{lsvS} \in {\mathbb{R}}^{d_{S}}$ by
\begin{equation}
    \textbf{lsvS}= \sum_{j=1}^{k} {\alpha_j \cdot \textbf{ls}_j}
\label{eq:lsvS}
\end{equation}
where $\alpha_j$ is the weight of $\textbf{ls}_j$ obtained by
\begin{equation}
    \alpha_{j} =\frac{\exp (\textbf{V}_a \cdot \tanh (\textbf{W}_a \cdot \textbf{ls}_{j} + \textbf{b}_a))}{\sum_{p=1}^{{k}} \exp (\textbf{V}_a \cdot \tanh (\textbf{W}_a \cdot \textbf{ls}_{p} + \textbf{b}_a))}
\end{equation}
where $\textbf{W}_a \in \mathbb{R}^{ d_{a} \times d_S }$ and $\textbf{V}_a \in \mathbb{R}^{ 1 \times d_a }$ are fully connected matrices, \textbf{b}$_a \in \mathbb{R}^{d_a}$ is the bias vector and $d_a$ is the dimension of the attention layer.

By this process, we obtain two latent state vectors, i.e., $\textbf{lsvI}$ and $\textbf{lsvS}$, which contain semantic information from prediction results and help the following predicting process.

\subsection{Dialogue History Representation}
In this section, we introduce DHR in detail, which utilizes the semantic information contained in the dialogue history efficiently.

Dialogue history contains much semantic information and can affect the prediction of the current utterance. Besides utterances, the predicted results in history also contain important semantics and are even more specific than utterances. Based on above, we design DHR that enable single-turn SLU models obtain contextual information contained in both utterance and predicted results of dialogue history.

\label{sec:DHR}

\begin{figure}[ht]
\centering
\tikzstyle{background1} = [rectangle, rounded corners=3mm,minimum width = 15cm, minimum height = 8.5cm, text centered, draw = black, fill = gray!8]
\tikzstyle{memory_list1} = [rectangle,rounded corners=1mm,minimum width = 4cm, minimum height = 0.6cm, text centered, draw = Peru, fill =Peach!15,thick]
\tikzstyle{memory_list2} = [rectangle,rounded corners=1mm,minimum width = 4cm, minimum height = 1.5cm, text centered, draw = Peru, fill =Peach!15,thick]
\tikzstyle{SLU_model} = [rectangle,rounded corners=1mm,minimum width = 1.6cm, minimum height = 0.6cm, text centered, draw = SeaGreen, fill = SeaGreen!15,thick]
\tikzstyle{input} = [rectangle,rounded corners=0mm,minimum width = 4.3cm, minimum height = 0.65cm, text centered, draw = Teal, fill = CornflowerBlue!15,thick]
\tikzstyle{output} = [rectangle,rounded corners=0mm,minimum width = 4.3cm, minimum height = 0.65cm, text centered, draw = Thistle, fill = Thistle!15,thick]

\begin{tikzpicture}[node distance = 0cm, scale=0.48]

\tikzstyle{every node}=[scale=0.48]
\node(background1_shadow)[background1,fill=gray!45,draw=gray!40]{};
\node(background1)[background1,below of=background1_shadow,xshift=-0.15cm,yshift=0.15cm]{};
\node(SLU_1)[SLU_model,below of=background1,xshift=0 cm,yshift=2.2cm]{RPFSLU};
\node(input1)[input,right of=SLU_1,xshift=-4.2cm,yshift=0cm,align=center]{\textbf{u}$_1$};
\node(memorylist1)[memory_list1,below of = input1,xshift=0cm,yshift=0.8cm,align=center]{};
\node()[right of=memorylist1,xshift=0 cm,yshift=0.5cm]{Memory List};
\node(output1)[output,right of=SLU_1,xshift=4.2cm,yshift=0cm,align=center]{\textbf{resI}$_1$,\textbf{resS}$_1$};
\node(memorylist2)[memory_list1,below of = input1,xshift=0cm,yshift=-1.25cm,align=center]{$<$ \textbf{u}$_1$,\textbf{resI}$_1$,\textbf{resS}$_1$ $>$};
\node()[right of=memorylist2,xshift=0 cm,yshift=0.5cm]{Memory List};

\node(omit_slu1)[below of=SLU_1,xshift=0 cm,yshift=-2.2cm]{...};
\node(omit_input1)[below of=input1,xshift=0cm,yshift=-2.2cm,align=center]{...};
\node()[below of=output1,xshift=0cm,yshift=-2.2cm,align=center]{...};

\node(SLU_2)[SLU_model,below of=omit_slu1,xshift=0 cm,yshift=-2.7cm]{RPFSLU};
\node(input2)[input,right of=SLU_2,xshift=-4.2cm,yshift=0cm,align=center]{\textbf{u}$_T$};
\node(memorylist3)[memory_list2,below of = input2,xshift=0cm,yshift=1.25cm,align=center]{$<$ \textbf{u}$_1$,\textbf{resI}$_1$,\textbf{resS}$_1$ $>$\\...\\$<$ \textbf{u}$_{T-1}$,\textbf{resI}$_{T-1}$,\textbf{resS}$_{T-1}$ $>$};
\node()[right of=memorylist3,xshift=0 cm,yshift=1cm]{Memory List};
\node(output2)[output,right of=SLU_2,xshift=4.2cm,yshift=0cm,align=center]{\textbf{resI}$_T$,\textbf{resS}$_T$};

\node(omit_slu2)[below of=SLU_2,xshift=0 cm,yshift=-1cm]{...};

\node(p1)[right of=memorylist1,xshift=2.5 cm,yshift=-0cm]{};
\node(p2)[right of=SLU_1,xshift=-1.6 cm,yshift=0.1cm]{};
\draw[-latex,densely dashed] (memorylist1.east) .. controls (p1) and (p2) .. (SLU_1.west); 
\draw[-latex,thick](input1.east) -- (SLU_1.west);
\draw[-latex,thick](SLU_1.east) -- (output1.west);
\node(p3)[below of=output1,xshift=0 cm,yshift=-1.115cm]{};
\draw[-,densely dashed](output1.south) -- (p3.south);
\draw[-latex,densely dashed](p3.south) -- (memorylist2.east);

\node(p4)[right of=memorylist3,xshift=2.5 cm,yshift=-0cm]{};
\node(p5)[right of=SLU_2,xshift=-1.6 cm,yshift=0.1cm]{};
\draw[-latex,densely dashed] (memorylist3.east) .. controls (p4) and (p5) .. (SLU_2.west); 
\draw[-latex,thick](input2.east) -- (SLU_2.west);
\draw[-latex,thick](SLU_2.east) -- (output2.west);


\end{tikzpicture}
\caption{ An illustration of the Memory List. }
\label{fig:memorylist}
\end{figure}
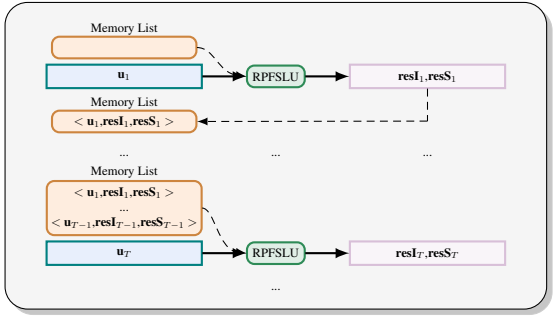

To record dialogue history, we employ memory lists to storage historical utterance and predicted results. As shown in Figure \ref{fig:memorylist}, the memory lists are empty initially and updated after each turn of dialogue.
After $T-1$ turn of dialogues, we have saved $T-1$ turn of the utterances, predicted ID results and SF results in memory lists 
$\textbf{M}= ( <\textbf{u}_1, \textbf{resI}_1, \textbf{resS}_1 >,..., <\textbf{u}_{T-1}, \textbf{resI}_{T-1}, \textbf{resS}_{T-1}> )$.

In $T$-th turn of dialogue, we first map all the historical results into the latent space by result representation mechanism and obtain the historical latent state vectors $\textbf{LSVI}=\{\textbf{lsvI}_1,...,\textbf{lsvI}_{T-1}\}$ for ID and $\textbf{LSVS}=\{\textbf{lsvS}_1,...,\textbf{lsvS}_{T-1}\}$ for SF.

At the same time, for each utterance $\textbf{u}_t$ from the memory list, we calculate its sentence-level semantic vector $\textbf{he}_t$ by
\begin{align}
    \textbf{e}^t &= Embedding(\textbf{u}_t)\\
    \textbf{he}_t &= BiGRU(\textbf{e}^t)
\end{align}
where $BiGRU$ is a Bi-directional GRU layer. 
We also calculate the semantic vector $\textbf{he}_T$ for the current utterance $\textbf{u}_T$ by the same way.

Actually, the historical dialogue with higher semantic similarity to the current dialogue should have a greater influence. Therefore, we calculate the weight of each historical dialogue $W_h=\{w_1,...,w_{T-1}\}$ by
\begin{align}
    w_{t} &= \frac{\exp (\textbf{he}_T^\top \cdot \textbf{he}_t)}{\sum_{j=1}^{T-1} \exp (\textbf{he}_T^\top \cdot \textbf{he}_j)}
\end{align}

Then, we calculate the weighted latent states vectors of historical ID and SF results by
\begin{align}
    \textbf{lsvI}_H &= \sum_{i=1}^{T-1} { w_t \cdot \textbf{lsvI}_t } \\
    \textbf{lsvS}_H &= \sum_{i=1}^{T-1} { w_t \cdot \textbf{lsvS}_t }
\end{align}

To incorporate contextual information into the basic model, we merge the historical latent state vectors with the current word embedding $\textbf{e}^T=\{\textbf{e}_1^T, ... , \textbf{e}_{k}^T  \}$ and obtain a new embedding $\textbf{e}^H=\{\textbf{e}_1^H, ... , \textbf{e}^H_{k} \}$ by
\begin{align}
    \textbf{e}^H_j &= \textbf{W}^H \cdot (\textbf{e}^T_j \oplus \textbf{lsvI}_H \oplus \textbf{lsvS}_H) + \textbf{b}^H
\label{eq:res2S}
\end{align}
where $\oplus$ is the concatenation operation, $\textbf{W}^H \in \mathbb{R}^{  d_w \times (d_{w}+d_I+d_S)}$ is a fully connected matrix, and $\textbf{b}^H \in \mathbb{R}^{ d_w }$ is the bias vector. 

$\textbf{e}^H$ contains current utterance information together with the utterance, ID and SF information from dialogue history. 
Finally, we input $\textbf{e}^H$ into any suitable SLU model and achieve the prediction results for the current utterance.


\subsection{Result-based Bi-Feedback Network}
\label{sec:RBFN}
In this section, we introduce the RBFN in detail, which incorporates semantic information in prediction results to the basic model and gain more comprehensive prediction results.

In practice, RBFN contains three steps. First, RBFN receives the word embedding sequence $\textbf{e}^H$ from DHR. Utilizing $\textbf{e}^H$, RBFN employs the basic model to achieve the first-round ID result distribution $\textbf{resI}^1$ and the first-round SF results distribution $\textbf{resS}^1 = \{ \textbf{s}^1_1, ... , \textbf{s}^1_{k} \}$.


As we aforementioned, in SLU, the prediction result of ID can affect the labeling of SF, while the results of the SF can verify the prediction of ID (so-called Bi-Feedback). Motivated by this, for the second step of RBFN, we obtain the latent states vector of the first-round results $\textbf{lsvI}$ and $\textbf{lsvS}$ by result representation mechanism. 
Then we merge the latent state vectors $\textbf{lsvI}$ and $\textbf{lsvS}$ with word embedding $\textbf{e}^H$ for the second-round prediction or verification, respectively.

Specifically, for the information of ID result, we first merge $\textbf{lsvI}$ with the utterance word embedding $\textbf{e}^H$ and obtain the ID-result based embedding sequence $\textbf{e}^I=\{ \textbf{e}^I_1,...,\textbf{e}^I_{k} \in {\mathbb{R}}^{d_w} \}$ by
\begin{align}
    \textbf{e}^I_j &= \textbf{W}^{I} \cdot (\textbf{lsvI} \oplus \textbf{e}^H_j) + \textbf{b}^I
\end{align}
where $\oplus$ is the concatenation operation, $\textbf{W}^I \in \mathbb{R}^{d_w \times (d_{w}+d_I)}$ is a fully connected matrix, and $\textbf{b}^I \in \mathbb{R}^{d_w}$ is the bias vector.

Similar as ID, for the information of SF result, we also merge $\textbf{lsvS}$ with the utterance word embedding $\textbf{e}^H$ and obtain the SF-result based embedding sequence $\textbf{e}^S=\{ \textbf{e}^S_1,...,\textbf{e}^S_{k} \in {\mathbb{R}}^{d_w} \}$ by
\begin{align}
    \textbf{e}^S_j &= \textbf{W}^{S} \cdot (\textbf{lsvS} \oplus \textbf{e}^H_j) + \textbf{b}^S
\end{align}
where $\textbf{W}^S \in \mathbb{R}^{  d_w \times (d_{w}+d_S)}$ is a fully connected matrix, and $\textbf{b}^S \in \mathbb{R}^{d_w}$ is the bias vector.

Subsequently, we employ the basic model again for the second-round prediction.
To utilize intent information for guiding SF process, we resend $\textbf{e}^I$ into the basic model to obtain the second round SF results $\textbf{resS}^2=\{ \textbf{s}^2_1, ... , \textbf{s}^2_{k} \}$.
To utilize slot information for verifying ID prediction, we resend $\textbf{e}^S$ into the basic model to obtain the second round ID results $\textbf{resI}^2$. Particularly, as $\textbf{resI}^2$ aims to verify $\textbf{resI}^1$, we replace $\mathrm{softmax}$ function with $\mathrm{sigmoid}$ function when outputting $\textbf{resI}^2$ from the basic model.

After obtaining two rounds of results, we merge them and calculate the final ID results $\textbf{resI}$ and SF results $\textbf{resS}=\{ \textbf{s}_1,...,\textbf{s}_{k} \}$ by
\begin{align}
    \textbf{resI} &= \textbf{resI}^1 \otimes \textbf{resI}^2\\
    \textbf{resS} &= \textbf{resS}^1 \otimes \textbf{resS}^2
\end{align}
where $\otimes$ is the element-wise product.

We finally normalize $\textbf{resI}$ and $\textbf{resS}$ by 
$f(x) = x_i / \sum x_j$.



\begin{table*}[t]
\centering
\resizebox{0.84\textwidth}{!}{
\begin{tabular}{l|ccc|ccc|ccc}
\hline
\multirow{2}{*}{\textbf{Model}} & \multicolumn{3}{c|}{\textbf{Basic Model}} & \multicolumn{3}{c}{\textbf{with DHR}} & \multicolumn{3}{c}{\textbf{with RPFSLU}}\\ 
& Intent(Acc.)   & Slot(F1)    & Overall(Acc.) & Intent(Acc.)   & Slot(F1)    & Overall(Acc.)  & \multicolumn{1}{l}{Intent(Acc.)} & \multicolumn{1}{l}{Slot(F1)} & \multicolumn{1}{l}{Overall(Acc.)} \\ \hline
\multirow{1}{*}{MemNet}         & 97.6  & 76.8  & 73.4  & -   & -     & - & 97.8(+0.2)     & 77.6(+0.8)  & 74.3(+1.1)\\\hline
\multirow{1}{*}{SDEN}           & 97.7*  & 76.9  & 75.3  & -   & -     & -& 97.8(+0.1)     & 77.4(+0.5)  & 76.0(+0.7)\\\hline
\multirow{1}{*}{SDEN+}          & 97.1  & 77.1*  & 75.7*  & -   & -     & -& 97.7(+0.6)     & 77.8(+0.7)  & 76.2(+0.5)\\\hline
\multirow{1}{*}{1-layer BiLSTM} & 97.1  & 75.9  & 74.3  & 98.0(+0.9)  & 77.6(+1.7) & 75.8(+1.5) & 98.8(+1.7)  & 78.5(+2.6) & 76.5(+2.2)\\\hline
\multirow{1}{*}{2-layers BiLSTM}& 97.3  & 76.5  & 75.3  & 98.5(+1.2)  & 78.0(+1.5) & 76.2(+0.9) & 98.7(+1.4)  & 78.3(+1.8) & 76.3(+1.0) \\\hline
\multirow{1}{*}{Joint Seq.}     & 97.2  & 76.1  & 75.1  & 98.1(+0.9)  & 78.1(+2.0) & 76.3(+1.2) & 98.6(+1.4)  & 78.2(+2.1) & 75.8(+0.7)\\\hline
\multirow{1}{*}{Bi-model}       & 96.7  & 75.1  & 72.2  & 98.5(+1.8)  & 77.7(+2.6) & 75.7(+3.5) & \textbf{98.9}(\textbf{+2.2})  & 79.1(+4.0) & \textbf{77.0}(\textbf{+4.8})\\\hline
{Stack-Propagation}             & 97.1  & 75.7  & 72.8  & 98.0(+0.9)  & 79.3(+3.6) & 76.6(+3.8) & 98.3(+1.2)  & \textbf{80.0}(\textbf{+4.3}) & 76.9(+4.1)\\\hline
\end{tabular}%
}
\caption{SLU performance on the KVRET dataset.}
\label{res2}
\end{table*}

\section{Experiments}
\subsection{Experimental Settings}
\label{4.1}
To evaluate the effectiveness of RPFSLU, we conduct experiments on a multi-turn dataset: KVRET \cite{kvret}.
The dataset consists of 3,031 multi-turn dialogues, where 2,425 dialogues in the training set, 302 in the validation set and 304 in the test set.
We utilize the validation set to choose hyper-parameters and evaluate the baseline models with our framework on the testing set. For each model, we conduct experiments for 50 times and select their best result.

In the training process, we set all hyper-parameters used in the basic model (e.g., the batch size, epochs, optimizer, learning rate, etc.) as their original paper. For hyper-parameters used in RPFSLU,
we set intent embedding size $d_I$ as 8, slot embedding size $d_S$ as 32, and set the word embedding size $d_w$ the same as the basic model.
The dimension of the attention layer $d_a$ is set as 64.


\subsection{Baselines}
\label{4.2}

To evaluate the performance of RPFSLU on multi-turn SLU tasks, we select the following existing multi-turn SLU models as benchmark.
(1) MemNet \cite{End-to-end2016}: A model with attention-based memory retrieval.
(2) SDEN \cite{sequentialSlu2017}: A model with sequential encoder based memory retrieval.
(3) SDEN$^+$ \cite{memory2019}: An advanced version of SDEN using different RNN structure in decoder.
We then set the following models as basic models of RPFSLU, which contain both simple SLU models such as 1-layer BiLSTM, and the state-of-the-art model Stack-Propagation to indicate the applicability of our framework:
\begin{itemize}
\item 1-layer BiLSTM: A single-layer Bidrectional LSTM model.
\item 2-layers BiLSTM: A two layers Bidrectional LSTM model.
\item Joint Seq \cite{Multi-domainJoint2016}: An GRU based model with multi-task modeling approach.
\item Bi-model \cite{birnn2018}: An RNN based encoder-decoder model considering the intent and slot ﬁlling cross-impact to each other.
\item Stack-Propagation \cite{qin2019stack}: An RNN based joint model with stack-propagation framework and token-level ID, which is the state-of-the-art of the joint model. 
%
\end{itemize}

\subsection{Results and Analysis}
\label{4.3}
Following \cite{memory2019}, we adopt accuracy for ID, F1 score for SF, and overall accuracy for whole sentence accuracy to evaluate the performance of each model.

\textbf{Results analysis:} Experimental results on the multi-turn SLU dataset KVRET are shown in Table \ref{res2}, where all SLU models in the experiment are benefited from our framework. For instance, via RPFSLU, the state-of-the-art model Stack-Propagation enhance the performance on each task by 1.2\%(ID), 4.3\%(SF), and 4.1\%(Overall) comparing with its original model. Among all models, Stack-Propagation achieves the best performance in SF while Bi-model achieves the best performance in ID and Overall accuracy. Comparing with the original model, Bi-model also achieves the largest margin of improvement when utilizing our framework.

Notably, all single-turn basic models in baselines perform worse than the benchmark without our framework, especially for SF. When utilizing RPFSLU, even a simple model (e.g., 1-layer BiLSTM) outperformed all benchmarks in all SLU tasks. Comparing 1-layer BiLSTM with SDEN+, our framework improves the performance by 1.7\%(ID), 1.4\%(SF), and 0.8\%(Overall). As for the state-of-the-art model Stack-Propagation, it enhances the performance by 1.2\%(ID), 2.9\%(SF), and 1.2\%(Overall).

\textbf{Ablation study:} 
Results show that even when DHR is activated alone, all models achieve a visible improvement. We further analyze the improvement of the state-of-the-art model. When utilizing DHR alone, Stack-Propagation enhances the performance by 0.9\%(ID), 3.6\%(SF), and 3.8\%(Overall) comparing with the original model. Then, we analyze the improvement of the simplest model. When utilizing DHR alone, 1-layer BiLSTM has already improved the performance by 0.9\%(ID), 2.2\%(SF), and 0.9\%(Overall) comparing with SDEN+.
This phenomenon indicates that our DHR is a better approach to obtain historical information than the methods used in those benchmarks. 
Meanwhile, as shown in Table \ref{res2}, although current multi-turn models cannot utilize DHR, they can benefit from the RPFSLU via RBFN module.
We also conduct experiments on two single-turn datasets to further verify the effectiveness of RBFN. These experimental results are shown in the appendix due to the limited space.



Above all, the experiment results indicate that RPFSLU is a portable and effective framework with high applicability, which has a significant improvement in multi-turn SLU tasks.
We attribute this to directly utilizing the prediction results since the results containing abundant semantic information. Combining the latent state of results with current utterance embedding makes the semantic information fully used and leads to well-improved performance. 
\section{Conclusion}
In this paper, we propose a Result-based Portable Framework for SLU, which allows most existing single-turn SLU models to take full advantage of contextual information and be able to deal with the multi-turn SLU tasks without changing their own structure.
Furthermore, we focus on the impact of prediction results on SLU tasks. Both the predicted results from dialogue history and the interaction between ID and SF in every single turn are utilized efficiently to improve the performance of existing SLU models without using any extra labeled data.
Experimental results indicate that RPFSLU benefits most SLU models on all tasks. 

\section{Acknowledgements}
This work is supported by Guangdong Key Lab of AI and Multi-modal Data Processing, Chinese National Research Fund (NSFC) Project No. 61872239; BNU-UIC Institute of Artificial Intelligence and Future Networks funded by Beijing Normal University (Zhuhai) and AI-DS Research Hub, BNU-HKBU United International College (UIC), Zhuhai, Guangdong, China.

\bibliographystyle{IEEEbib}
\bibliography{ICME2021.bib}

\begin{thebibliography}{10}

\bibitem{SLU2011}
Gokhan Tur and Renato De~Mori,
\newblock {\em Spoken language understanding: Systems for extracting semantic
  information from speech},
\newblock John Wiley \& Sons, 2011.

\bibitem{bi-dictional2019}
E~Haihong, Peiqing Niu, Zhongfu Chen, and Meina Song,
\newblock ``A novel bi-directional interrelated model for joint intent
  detection and slot filling,''
\newblock in {\em ACL}, 2019, pp. 5467--5471.

\bibitem{End-to-end2016}
Antoine Bordes, Y{-}Lan Boureau, and Jason Weston,
\newblock ``Learning end-to-end goal-oriented dialog,''
\newblock in {\em ICLR}, 2017.

\bibitem{sequentialSlu2017}
Ankur Bapna, Gokhan T{\"u}r, Dilek Hakkani-T{\"u}r, and Larry Heck,
\newblock ``Sequential dialogue context modeling for spoken language
  understanding,''
\newblock in {\em SIGDIAL}, 2017, pp. 103--114.

\bibitem{memory2019}
He~Bai, Yu~Zhou, Jiajun Zhang, and Chengqing Zong,
\newblock ``Memory consolidation for contextual spoken language understanding
  with dialogue logistic inference,''
\newblock in {\em ACL}, 2019, pp. 5448--5453.

\bibitem{Slot-gated2018}
Chih-Wen Goo, Guang Gao, Yun-Kai Hsu, Chih-Li Huo, Tsung-Chieh Chen, Keng-Wei
  Hsu, and Yun-Nung Chen,
\newblock ``Slot-gated modeling for joint slot filling and intent prediction,''
\newblock in {\em NAACL-HLT}, 2018, pp. 753--757.

\bibitem{JointCapsule2019}
Chenwei Zhang, Yaliang Li, Nan Du, Wei Fan, and Philip~S Yu,
\newblock ``Joint slot filling and intent detection via capsule neural
  networks,''
\newblock in {\em ACL}, 2019, pp. 5259--5267.

\bibitem{qin2019stack}
Libo Qin, Wanxiang Che, Yangming Li, Haoyang Wen, and Ting Liu,
\newblock ``A stack-propagation framework with token-level intent detection for
  spoken language understanding,''
\newblock in {\em EMNLP-IJCNLP}, 2019, pp. 2078--2087.

\bibitem{qin2021survey}
Libo Qin, Tianbao Xie, Wanxiang Che, and Ting Liu,
\newblock ``A survey on spoken language understanding: Recent advances and new
  frontiers,''
\newblock {\em arXiv preprint arXiv:2103.03095}, 2021.

\bibitem{yao2014lstm}
Kaisheng Yao, Baolin Peng, Yu~Zhang, Dong Yu, Geoffrey Zweig, and Yangyang Shi,
\newblock ``Spoken language understanding using long short-term memory neural
  networks,''
\newblock in {\em SLT}. IEEE, 2014, pp. 189--194.

\bibitem{kurata2016lstm}
Gakuto Kurata, Bing Xiang, Bowen Zhou, and Mo~Yu,
\newblock ``Leveraging sentence-level information with encoder {LSTM} for
  semantic slot filling,''
\newblock in {\em EMNLP}, 2016, pp. 2077--2083.

\bibitem{JointModel2016}
Xiaodong Zhang and Houfeng Wang,
\newblock ``A joint model of intent determination and slot filling for spoken
  language understanding.,''
\newblock in {\em IJCAI}, 2016, pp. 2993--2999.

\bibitem{chen2016joint}
Yun-Nung Chen, Dilek Hakanni-T{\"u}r, Gokhan Tur, Asli Celikyilmaz, Jianfeng
  Guo, and Li~Deng,
\newblock ``Syntax or semantics? knowledge-guided joint semantic frame
  parsing,''
\newblock in {\em SLT}. IEEE, 2016, pp. 348--355.

\bibitem{attention-basedRnn2016}
Bing Liu and Ian Lane,
\newblock ``Attention-based recurrent neural network models for joint intent
  detection and slot filling,''
\newblock in {\em Interspeech}, 2016, pp. 685--689.

\bibitem{Multi-domainJoint2016}
Dilek Hakkani-T{\"u}r, G{\"o}khan T{\"u}r, Asli Celikyilmaz, Yun-Nung Chen,
  Jianfeng Gao, Li~Deng, and Ye-Yi Wang,
\newblock ``Multi-domain joint semantic frame parsing using bi-directional
  rnn-lstm.,''
\newblock in {\em Interspeech}, 2016, pp. 715--719.

\bibitem{Multi-TaskNetworks2019}
Shiva Pentyala, Mengwen Liu, and Markus Dreyer,
\newblock ``Multi-task networks with universe, group, and task feature
  learning,''
\newblock in {\em ACL}, 2019, pp. 820--830.

\bibitem{qin2020co}
Libo Qin, Tailu Liu, Binging~Kang Wanxiang~Che, and Ting Liu,
\newblock ``A co-interactive transformer for joint slot filling and intent
  detection,''
\newblock in {\em ICASSP}, 2021.

\bibitem{qin2020dcr}
Libo Qin, Wanxiang Che, Yangming Li, Mingheng Ni, and Ting Liu,
\newblock ``Dcr-net: A deep co-interactive relation network for joint dialog
  act recognition and sentiment classification,''
\newblock in {\em AAAI}.

\bibitem{qin2021knowing}
Libo Qin, Wanxiang Che, Minheng Ni, Yangming Li, and Ting Liu,
\newblock ``Knowing where to leverage: Context-aware graph convolution network
  with an adaptive fusion layer for contextual spoken language understanding,''
\newblock {\em TASLP}, 2021.

\bibitem{liu2019cm}
Yijin Liu, Fandong Meng, Jinchao Zhang, Jie Zhou, Yufeng Chen, and Jinan Xu,
\newblock ``Cm-net: A novel collaborative memory network for spoken language
  understanding,''
\newblock in {\em EMNLP-IJCNLP}, 2019, pp. 1050--1059.

\bibitem{BERT}
Jacob Devlin, Ming-Wei Chang, Kenton Lee, and Kristina Toutanova,
\newblock ``Bert: Pre-training of deep bidirectional transformers for language
  understanding,''
\newblock in {\em NAACL}, 2019.

\bibitem{chen2019bert}
Qian Chen, Zhu Zhuo, and Wen Wang,
\newblock ``Bert for joint intent classification and slot filling,''
\newblock {\em arXiv preprint arXiv:1902.10909}, 2019.

\bibitem{birnn2018}
Yu~Wang, Yilin Shen, and Hongxia Jin,
\newblock ``A bi-model based {RNN} semantic frame parsing model for intent
  detection and slot filling,''
\newblock in {\em NAACL-HLT}, 2018, pp. 309--314.

\bibitem{yang2019}
Wenmian Yang, Weijia Jia, XIaojie Zhou, and Yutao Luo,
\newblock ``Legal judgment prediction via multi-perspective bi-feedback
  network,''
\newblock in {\em IJCAI}, 2019, pp. 4085--4091.

\bibitem{kvret}
Mihail Eric, Lakshmi Krishnan, Francois Charette, and Christopher~D. Manning,
\newblock ``Key-value retrieval networks for task-oriented dialogue,''
\newblock in {\em SIGDIAL}, 2017, pp. 37--49.

\end{thebibliography}
\end{spacing}
\end{document}